\pdfoutput=1

\documentclass[11pt]{article}

\usepackage[preprint]{acl}

\usepackage{times}
\usepackage{latexsym}
\usepackage{tcolorbox}
\usepackage{times}
\usepackage{latexsym}
\usepackage{graphicx,tabularx}
\usepackage{multirow}
\usepackage{multicol}
\usepackage{amsmath}

\usepackage{amsfonts}
\usepackage{booktabs}
\usepackage{float}
\usepackage{algorithm}
\usepackage{algpseudocode}
\usepackage{xcolor}
\usepackage{makecell} 
\usepackage[normalem]{ulem}
\usepackage{array}
\usepackage{arydshln}
\usepackage{booktabs}
\usepackage{multirow}

\usepackage[T1]{fontenc}

\usepackage[utf8]{inputenc}

\usepackage{microtype}

\usepackage{inconsolata}

\usepackage{graphicx}

\title{Mitigating Hallucinations in Large Vision-Language Models via Summary-Guided Decoding}

\author{%
Kyungmin Min$^1$ \quad Minbeom Kim$^1$ \quad Kang-il Lee$^2$ \\
\textbf{Dongryeol Lee}$^2$ \quad  \textbf{Kyomin Jung}$^{1,2}\thanks{Corresponding authors.}$\\
$^1$IPAI, Seoul National University\\ $^2$Dept. of ECE, Seoul National University  \\
\texttt{\{kyungmin97,minbeomkim,4bkang,drl123,kjung\}@snu.ac.kr}
}

\begin{document}
\maketitle

\definecolor{pastelred}{RGB}{255, 0, 0}

\begin{abstract}



Large Vision-Language Models (LVLMs) demonstrate impressive capabilities in generating detailed and coherent responses from visual inputs.
However, they are prone to generate hallucinations due to an over-reliance on language priors. 
To address this issue, we investigate the language priors in LVLMs and make two key observations: (1) Even when predicting the tokens associated with image-related part-of-speech (POS), models increasingly rely on linguistic priors as the token sequences grow, thereby amplifying hallucinations. (2) Methods that directly calibrate LVLM's output distribution to mitigate language priors can lead to a degradation in text quality or even exacerbate hallucinations.
Based on these findings, we propose a novel method, \textbf{Sum}mary-\textbf{G}uided \textbf{D}ecoding \textbf{(SumGD)}. This method naturally encourages the model to focus more on image information by reducing the text context through summaries, while controlling only the image-related POS tokens to maintain text quality.
Through experiments, we demonstrate that SumGD achieves state-of-the-art performance on object hallucination benchmarks. 
Furthermore, in terms of the trade-off between precision and recall, SumGD achieves Pareto optimality among the existing methods.
Lastly, we observe that although existing methods struggle to balance the reduction of object hallucinations with maintaining text quality, SumGD demonstrates robustness in handling this challenge.


\end{abstract}

\section{Introduction}
\label{introduction}

\begin{figure}[t]
\centering
\includegraphics[trim=0 0 0 15, clip,width=1\linewidth]{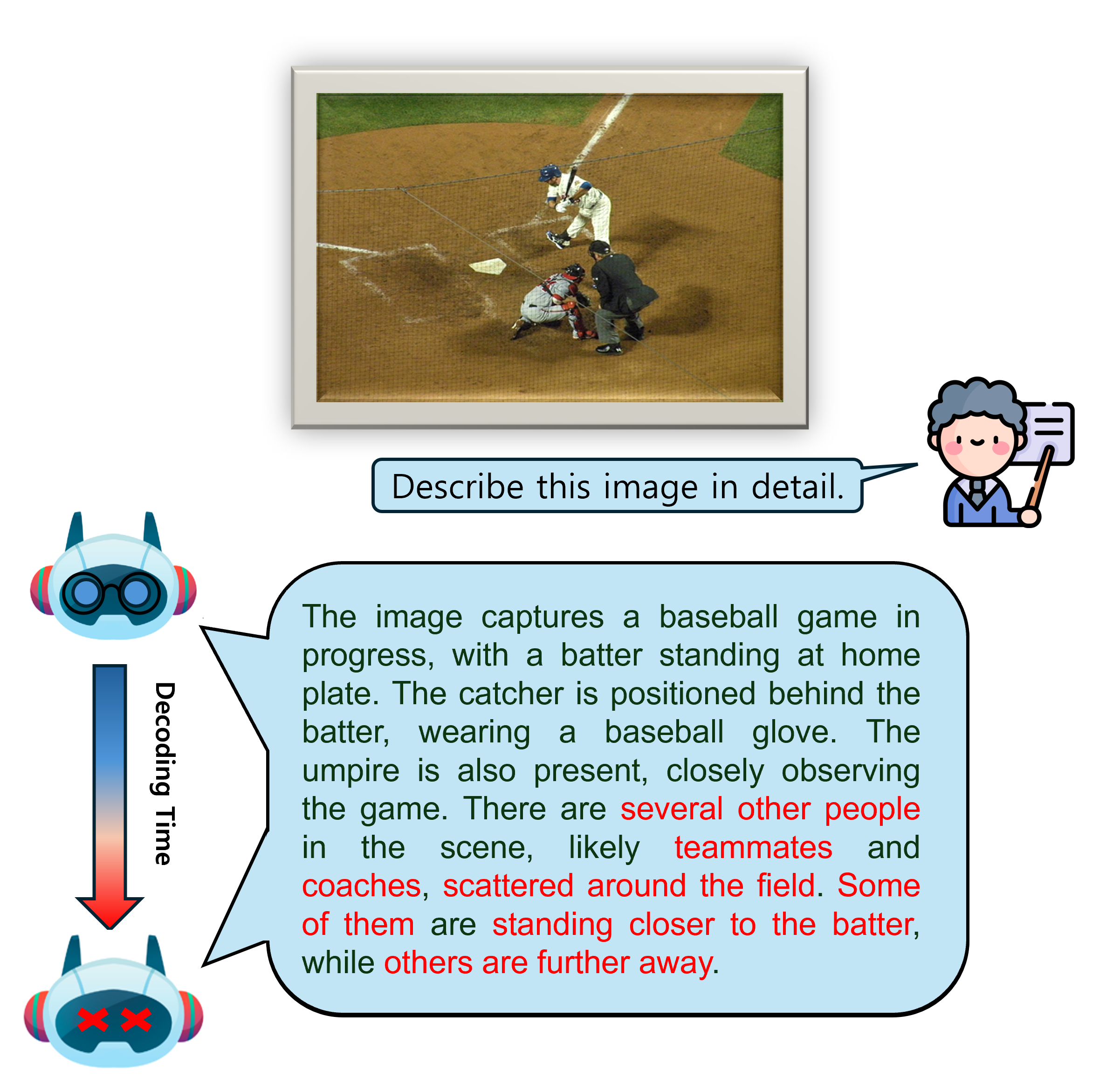}
\caption{An example of LVLMs' hallucination. LVLMs hallucinate due to their over-reliance on previously generated text. The \textcolor{red}{red} fonts represent the hallucinatory content.}
\label{intro_figure}
\vspace{-4mm}
\end{figure}

Large Vision-Language Models (LVLMs) have shown remarkable advancements by integrating the reasoning capabilities of Large Language Models (LLMs) to interpret visual knowledge~\citep{zhu2023minigpt4enhancingvisionlanguageunderstanding,dai2023instructblipgeneralpurposevisionlanguagemodels, liu2024improvedbaselinesvisualinstruction,li2023blip2bootstrappinglanguageimagepretraining}. 
Despite their significant utility, they suffer from a critical drawback known as \textit{object hallucination}, where the model generate responses that contradict the visual input~\cite{li-etal-2023-evaluating, liu2024survey}.
Recent studies have shown that this occurs because LVLMs rely too heavily on learned textual patterns, which referred as \textit{language priors}~\cite{zhou2024analyzingmitigatingobjecthallucination,liu2024mitigatinghallucinationlargemultimodal,jing2023faithscoreevaluatinghallucinationslarge,lee2024vlindbenchmeasuringlanguagepriors}.
This over-reliance on language priors tends to intensify when the model generates longer sequences or detailed descriptions~\cite{favero2024multimodalhallucinationcontrolvisual}, leading to frequent hallucinations as shown in Figure~\ref{intro_figure}.

In this paper, we 1) conduct the fundamental analysis of language priors in LVLMs (Section~\ref{total section2}), 2) analyze the limitations of existing methods for mitigating language priors and provide insights into potential solutions (Section~\ref{analysis language prior text}), and 3) propose a novel method that effectively reduces object hallucination while preserving text quality (Section~\ref{sec_3}).

First, we analyze language priors by examining the distance between the next-token probability distributions of LVLMs and LLMs, both conditioned on the same text sequence.
Breaking this down by part-of-speech (POS) types reveals a significant divergence for image-related POS tokens, such as NOUN (e.g., ``tree'') or ADJ (e.g., ``green''). Conversely, language-related POS tokens, such as AUX (e.g., ``is,'' ``will''), show nearly identical distributions.
These findings suggest that LVLMs still rely heavily on the same linguistic structures as LLMs, except when visual input is particularly relevant — such as when describing specific objects or attributes. In other words, LVLMs incorporate visual information within a linguistic framework that is very similar to that of LLMs.

Problematically, we discover that even for these image-related POS tokens, the distributional distance rapidly decreases as the number of generated tokens increases.
In other words, even when visual information is necessary, LVLMs tend to focus more on textual information, leading to frequent occurrences of object hallucination. We identify this phenomenon as an over-reliance on language priors.

Next, we examine the limitations of contrastive decoding, a promising methods for mitigating hallucinations~\cite{favero2024multimodalhallucinationcontrolvisual,wang2024mitigatinghallucinationslargevisionlanguage,leng2023mitigatingobjecthallucinationslarge,kim2024vacodevisualaugmentedcontrastive,zhu2024ibdalleviatinghallucinationslarge}. 
Our analysis reveals two primary issues: (1) The effort to reduce language priors through contrastive decoding can disrupt the natural distribution of language-related tokens, potentially degrading overall text quality.
(2) As token length increases, the model's reliance on language priors becomes more pronounced, leading the two output distributions being contrasted to become increasingly similar. This similarity reduces the effectiveness of contrastive decoding in steering the model towards an image-aligned distribution.
These findings suggest that reducing language priors may be more effectively achieved by integrating visual information naturally, with minimal intervention in the decoding process.

Building on these observations, we propose a novel method called Summary-Guided Decoding (SumGD).
Our approach employs a summarization technique that selectively retains essential information from previously generated sentences, encouraging LVLMs to more effectively incorporate image information.
To minimize unnecessary intervention for preserving text quality, the summarization is referenced only when predicting image-related POS tokens, which require image-specific details.


Our experimental results demonstrate that SumGD significantly outperforms all other decoding approaches in object hallucination benchmarks (e.g., up to +16.5\% in CHAIR$_S$ and +19\% in CHAIR$_I$) across various models and architecture sizes.
Additionally, SumGD demonstrates Pareto optimal performance, effectively balancing the reduction of object hallucinations with the preservation of high object recall. This balance becomes more pronounced as token length increases.
Finally, the results confirm that SumGD not only reduces object hallucinations but also preserves the overall text quality of LVLMs.\footnote{The code will be available at \url{https://github.com/andy9705/SumGD}}

Our contributions are summarized as follows:
\begin{itemize}
    \item We analyze how LVLMs tend to disregard image information and increasingly rely on language priors, based on the position and POS type of each token.

    \item Based on these findings, we propose Summary-Guided Decoding (SumGD). SumGD modifies next-token probabilities using summarized contexts, but only for image-related POS tokens. This approach aims to reflect image information while preserving LVLM's text quality as much as possible.
    \item SumGD demonstrates state-of-the-art performance in object hallucination benchmarks and achieves Pareto optimal across all methods in terms of the precision-recall trade-off. Additionally, SumGD preserves text quality almost entirely.
    
\end{itemize}

\section{Language Priors in LVLMs}
\label{total section2}
In this section, we systematically analyze the causes of language priors in LVLMs. 
Section~\S\ref{how to measure JSD} outlines the method for quantifying language priors. 
Section~\S\ref{motivation_pos_sec} provides an in-depth analysis of how language priors affect LVLMs based on part-of-speech (POS) types.
Section~\S\ref{2.3} analyzes the impact of increasing token length on language priors in LVLMs. 
We conduct this analysis on 5,000 MSCOCO~\cite{lin2015microsoftcococommonobjects} image descriptions generated using LLAVA 1.5 7B~\cite{liu2024improvedbaselinesvisualinstruction} (see Appendix~\ref{motivation_experiemtn_settings} for more details).

\subsection{How to measure language priors in LVLMs}
\label{how to measure JSD}
In LVLMs, language priors refer to the model’s over-reliance on learned textual patterns, where responses are generated based on these patterns without fully considering the provided image.
From this perspective, if the token distribution of a LVLM, which decodes using both text and images, becomes similar to that of a LLM, which relies solely on text for decoding, this could indicate an over-reliance on language priors.
Here, the \textit{LLM} refers to the state of the LVLM where the input image is not provided as a conditioning factor, with both models conditioned on the same text sequence.
Therefore, we measure language priors by examining the distributional distance between the next-token probabilities of LVLMs and LLMs, as described in \citet{favero2024multimodalhallucinationcontrolvisual}.
We employ Jensen-Shannon Divergence (JSD)~\cite{61115} to quantify this distance.

Formally, at each time step \( t \), the next token \( y_t \) is selected as:
\begin{equation}
y_{t} = \arg\max_{y \in V} \log p_{\theta}(y \mid I, T, y_{<t}),
\end{equation}
where \( \theta \) is the parameters of LVLMs, \( V \) is the vocabulary, \( I \) denotes the provided image, \( T \) represents the textual prompt (e.g., ``Please describe this image in detail.''), and \( y_{<t} \) denotes the sequence of generated tokens up to the time step (\( t-1 \)).

We define the distributional distance at each time step \( t \) as:
\begin{equation}\label{distance_t}
\operatorname{dist}_t = \mathrm{JSD}\left(p_{\theta}(\cdot \mid I, T, y_{<t}) \,\|\, p_{\theta}(\cdot \mid T, y_{<t})\right).
\end{equation}
A larger distance \( \operatorname{dist}_t \) suggests that the LVLM relies more on visual information for predictions, indicating a lower dependence on language priors. Conversely, a smaller distance implies that the model is generating responses primarily based on textual patterns.

\begin{figure}[t]
\vspace{-0.9cm}
  \centering
    \includegraphics[trim=0 0 0 20, clip,width=1\linewidth]{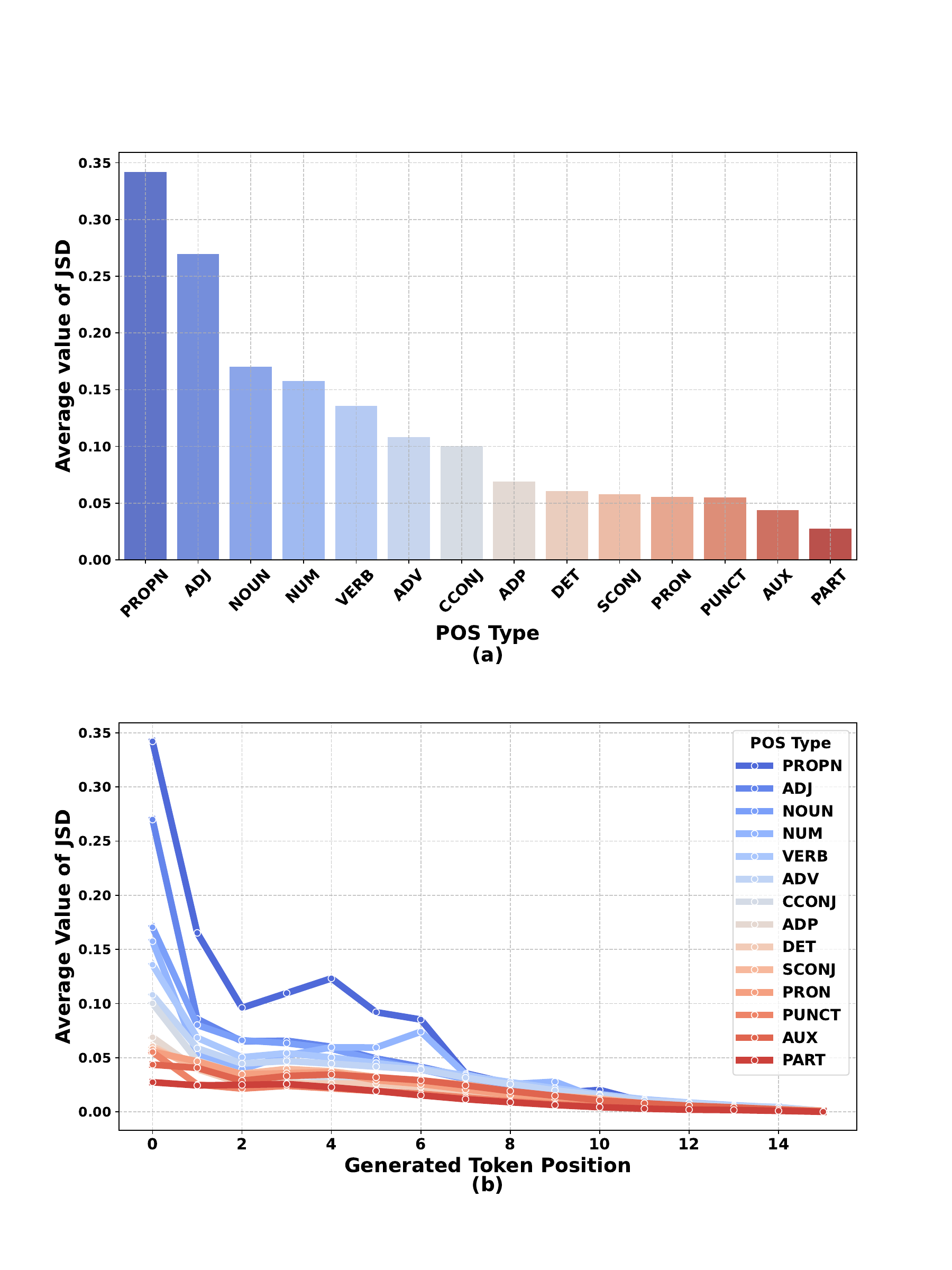}
    \vspace{-1.2cm}
\caption{
    \textbf{(Top)} 
The average JSD between the LVLM and the LLM for each POS category up to 32 tokens.
\textbf{(Bottom)} 
The average JSD between the LVLM and the LLM for each POS category across intervals, with each interval consisting of 32 tokens.
    }
    \label{motivation_jsd}
\vspace{-1mm}
\end{figure}

\begin{figure}[th]
  \centering
    \includegraphics[width=1\linewidth]{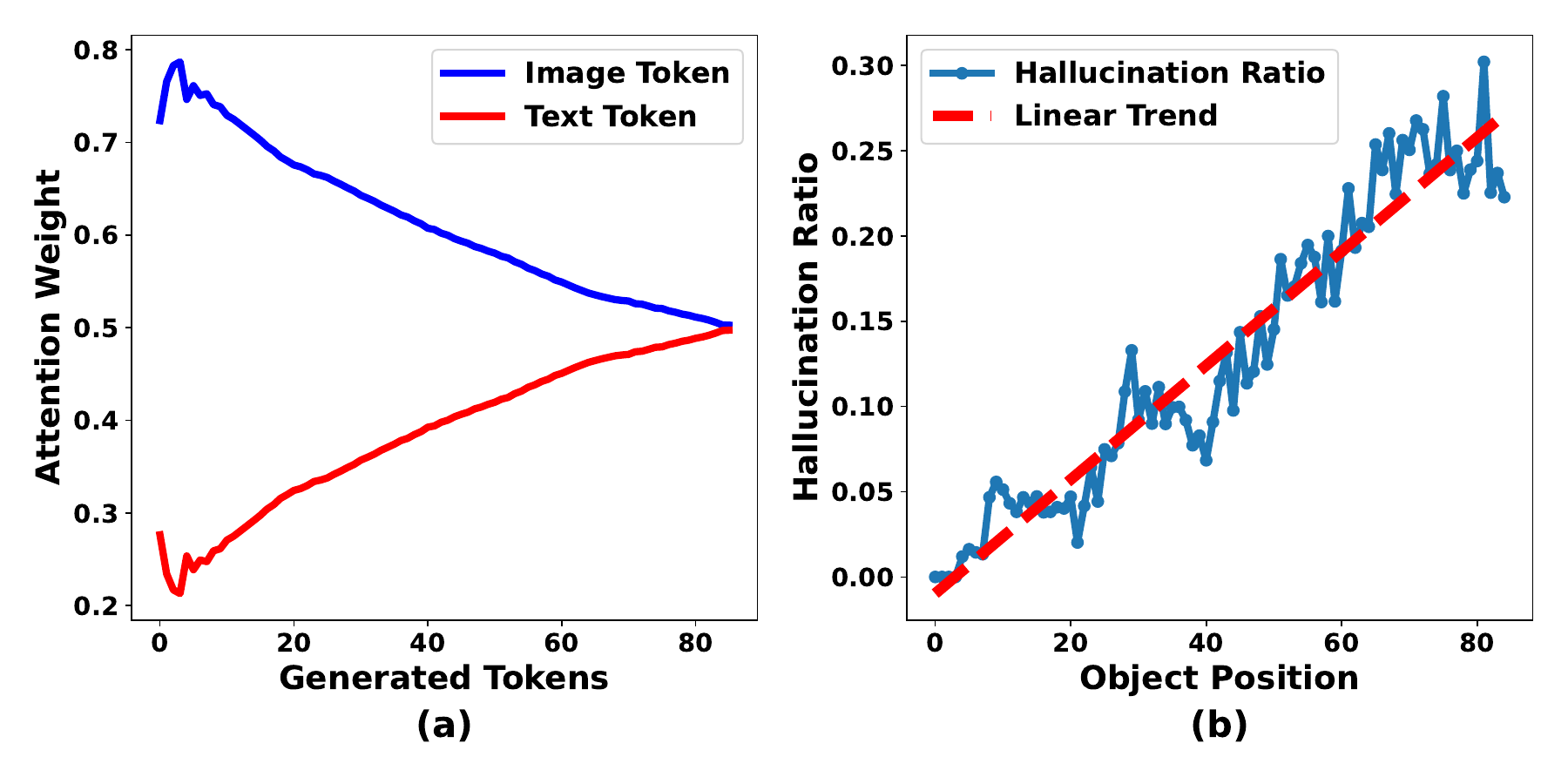}
    \vspace{-0.9cm}
\caption{
    \textbf{(Left)} Attention weights of image tokens and text tokens at each decoding step (or token length). \textbf{(Right)}  Object hallucination ratio at each generated token position.
    }
    \label{input_length_language_prior_2}
\vspace{-4mm}
\end{figure}

\subsection{Analysis of language priors by Part-of-Speech (POS) type}
\label{motivation_pos_sec}

\begin{figure*}[t]
\centering
\includegraphics[width=1.0\textwidth]{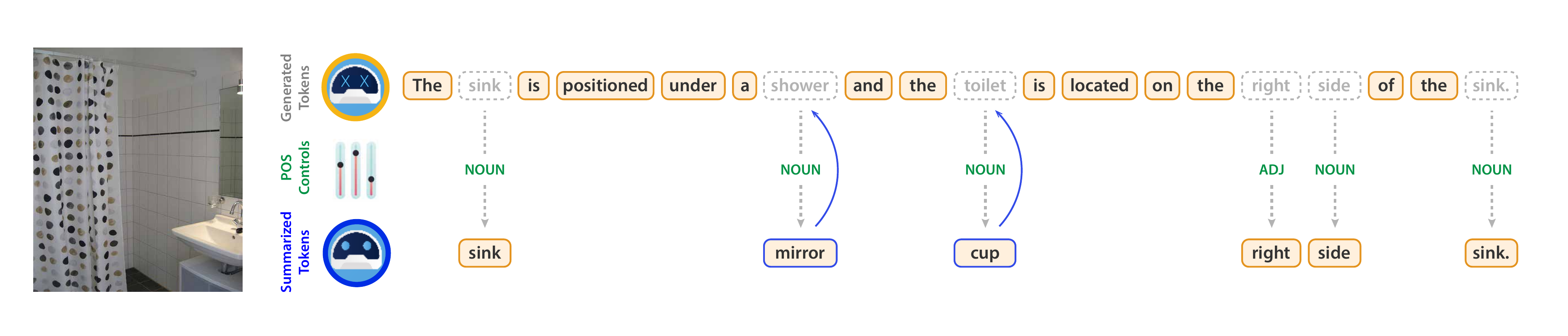} 
\caption{Illustration of our Summary-Guided Decoding.}
\label{pipeline}
\end{figure*}

We conduct an experiment to investigate whether LVLMs differ in their reliance on language priors based on the need for image information.
Specifically, we analyze this tendency by part-of-speech (POS) type, measuring the JSD at each decoding step and averaging the JSD values for each POS type\footnote{We utilized the Spacy model (en\_core\_web\_sm) for POS tagging} up to 32 tokens.

As shown in Figure~\ref{motivation_jsd} (a), we observe significant variation in divergence across different POS categories. POS categories such as PROPN (e.g., ``Biden'') and ADJ (e.g., ``red''), which related to visual information, exhibit higher divergence. 
In contrast, language-related POS types, like PART (e.g., ``not'', ``'s'') and AUX (e.g., ``are''), show much lower JSD. 
This indicates that LVLMs integrate visual information within a linguistic framework that is highly aligned with LLMs.

Another important observation, as shown in Figure~\ref{motivation_jsd} (b), is that even for image-related POS tokens (e.g., NOUN), the distributional distance decreases significantly as the token length increases.

This suggests that even when image information is required during decoding, models primarily rely on textual patterns.
In other words, token length (or input length) has a significant influence on how language priors are employed.

\subsection{Influence of Token Sequence Length on language priors}
\label{2.3}

We observe that as token sequences grow longer, the model becomes increasingly dependent on language priors in Section~\ref{motivation_pos_sec}.
To explore this effect further, we conduct a detailed analysis of how varying token lengths impact LVLMs, particularly in terms of how attention is distributed between image and text tokens, and the consequent impact on object hallucination.

First, we measure the attention weights assigned to image tokens and text tokens at each decoding step.
As shown in Figure~\ref{input_length_language_prior_2} (a), initially, LVLMs give sufficient attention to input image tokens when computing the next token. However, as the sentence grows longer, this attention becomes significantly shallower. In other words, when generating long sentences, we can observe that LVLMs tend to rely more on linguistic patterns rather than on visual information.
This observation provides additional insight into our earlier findings in Section~\ref{motivation_pos_sec}, where longer sequences were shown to reinforce the model's dependence on language priors.

Additionally, to assess the role of input length in hallucination, we evaluate the object hallucination ratio as a function of token length.
Figure~\ref{input_length_language_prior_2} (b) shows a clear correlation between input length and the likelihood of object hallucinations, indicating that longer text generation increases the chances of hallucination. 
We hypothesize that this phenomenon is driven by over-reliance on language priors, which amplifies hallucinations in LVLMs.

\section{Summary-Guided Decoding}
\label{sec_3}

Based on insights from Section~\ref{total section2},
we identify that an increase in input length results in greater reliance on language priors, thereby exacerbating hallucinations in LVLMs.
To address this, we present Summary-Guided Decoding (SumGD), a novel method for controlling the length of conditioning input during decoding.
In SumGD, we shorten the conditioning input by summarizing the previously generated text after each sentence completion.
This process preserves the critical context from earlier outputs while keeping the input concise.
The summarized text, combined with the image, serves as part of the conditioned input for generating the next sentence. This approach effectively reduces the input length, allowing the model to stay more focused on the provided image.

Using summarized inputs can reduce contextual information, which may cause discrepancies with the language patterns previously learned by the model. This can result in distributional shifts that weaken the model's language modeling capabilities, ultimately degrading the quality of the generated text.
To address this, we preserve the original distribution for tokens related to language modeling, while using SumGD to control only the image-related POS tokens.\footnote{As shown in Figure~\ref{motivation_jsd}, we selected PROPN, ADJ, NOUN, and NUM as image-related POS.} 
Our method is illustrated in Figure~\ref{pipeline}.

We introduce two variations of SumGD for summary model usage.
The first approach leverages the instruction-following capabilities inherent in LVLMs. 
By providing summary instructions directly to the LVLM, this method enables the model to perform SumGD without incurring additional memory costs.
However, a limitation of this approach is the increased computational burden, as the LVLM generates its summaries during the process.
To address these challenges, we distill the summarization capability into a smaller, more efficient model, Flan-T5-base~\cite{chung2022scalinginstructionfinetunedlanguagemodels} (see Appendix~\ref{distill_training} for details). This model significantly reduces computational overhead while maintaining the advantage of input length control. We report results for both \textbf{SumGD with Self-Summarization (SumGD-S)} and \textbf{SumGD with the Distilled-Flan-T5 model (SumGD-D)}, highlighting the trade-offs between efficiency and performance.

\section{Experiment}

\begin{table*}[t]
\centering
\resizebox{1\textwidth}{!}{%
\begin{tabular}{lccccccccccccccc} 
\toprule 
\multicolumn{1}{c}{\multirow{2}{*}{Method}} & \multicolumn{3}{c}{LLAVA-1.5 7B} & \multicolumn{3}{c}{InstructBLIP 7B} & \multicolumn{3}{c}{LLAVA-1.5 13B} & \multicolumn{3}{c}{InstructBLIP 13B} & \multicolumn{3}{c}{Average} \\
\cmidrule(r){2-4} \cmidrule(l){5-7} \cmidrule(l){8-10} \cmidrule(l){11-13} \cmidrule(l){14-16}
 & C$_S$ $\downarrow$ & C$_I$ $\downarrow$ & R $\uparrow$ & C$_S$ $\downarrow$ & C$_I$ $\downarrow$ & R $\uparrow$ & C$_S$ $\downarrow$ & C$_I$ $\downarrow$ & R $\uparrow$ & C$_S$ $\downarrow$ & C$_I$ $\downarrow$ & R $\uparrow$ & C$_S$ $\downarrow$ & C$_I$ $\downarrow$ & R $\uparrow$ \\
\midrule
Greedy       & 51.5 & 13.7 & \underline{79.1} & 49.0  & 15.6   & 72.7 & 43.5 & 12.2 & \underline{78.3} & 52.0 & 13.5 & 69.8 & 49.0 & 13.8 & \underline{75.0} \\
Nucleus      & 53.0 & 14.4 & 76.9 & 57.0 & 16.9 & 72.3 & 49.5 & 14.3 & 74.4 & 64.5 & 19.2 & 68.6 & 56.0 & 16.2 & 73.1 \\
\midrule
{\textit{\textbf{Beam search based (n=5)}}}       &&&&&&&&&&&&&&&  \\
Beam Search   & 47.5 & 12.5 & \textbf{79.2} & 45.5 & 13.1 & \textbf{74.1} & 43.5 & 12.0 & \underline{78.3} & 58.5 & 15.0 & 71.1 & 48.8 & 13.2 & \textbf{75.7} \\
OPERA  & 46.0 & 13.4 & 78.3 & \underline{43.0} & 13.0 & \underline{73.8} & \textbf{40.0} & 12.5 & 72.2 & \textbf{44.5} & 12.0 & 69.5 & 43.4 & 12.7 & 73.5 \\
\midrule
{\textit{\textbf{Contrastive Decoding}}}       &&&&&&&&&&&&&&&  \\
VCD          & 58.0 & 16.4 & 77.8 & 56.5 & 16.5  & 71.6 & 59.5 & 16.8 & \textbf{79.5} & 52.5 & 13.4 & \underline{71.2} & 56.6 & 15.8 & \underline{75.0} \\
ICD          & 45.5 & 13.4 & 77.2 & 60.5  & 17.8 & 68.9 & 47.5 & 13.0 & 77.3 & 66.0 & 19.3 & \textbf{72.2} & 54.9 & 15.9 & 73.9 \\
M3ID         & 44.5 & 12.0 & 76.1 & 68.0 & 18.0 & 71.6 & 45.0 & 11.9 & 77.8 & 78.0 & 20.8 & 67.8 & 58.9 & 15.7 & 73.3 \\
\midrule
\textbf{SumGD-D (Ours)} & \textbf{42.5} & \underline{11.8} & 77.8 & \textbf{42.5} & \underline{12.3} & 72.7 & 43.0 & \textbf{10.9} & 77.7 & \textbf{44.5} & \underline{11.6} & 69.2 & \textbf{43.1} & \underline{11.7} & 74.4 \\
\textbf{SumGD-S (Ours)} & \underline{43.0} & \textbf{11.1} & \underline{79.1} & 43.5 & \textbf{11.9} & 72.2 & \underline{41.5} & \underline{11.7} & 77.3 & \textbf{44.5} & \textbf{10.4} & 68.8 & \textbf{43.1} & \textbf{11.3} & 74.4 \\
\bottomrule
\end{tabular}%
}
\caption{Results on CHAIR Metric (\textit{max new tokens} is 512). The best performances are bolded, and the second-best are underlined. Denote CHAIR$_S$ as $C_S$, CHAIR$_I$ as $C_I$, and Recall as $R$. $n$ denotes the number of beams.}
\label{method_chair_table_2}
\vspace{-4mm}
\end{table*}

\subsection{Experiment settings}
\label{experiment_Settings}
\noindent\textbf{Datasets and Evaluation Metrics.}
We generate descriptions for 200 images from the MSCOCO 2014 validation dataset~\cite{lin2015microsoftcococommonobjects} prompted with \texttt{``Please describe this image in detail.''}~\cite{huang2024operaalleviatinghallucinationmultimodal}.
We employ the Caption Hallucination Assessment with Image Relevance (CHAIR)~\cite{rohrbach2019objecthallucinationimagecaptioning} for evaluating object hallucination. 
CHAIR consists of two variants: CHAIR$_I$, which calculates the percentage of hallucinated objects out of all objects mentioned in the caption, and CHAIR$_S$, which measures the percentage of captions that contain at least one hallucinated object.
Additionally, to complement the precision-based CHAIR metric, we introduce a Recall metric for a more detailed assessment.

{\small
$$\text{CHAIR}_{I} = \frac{ \vert\{ \text{hallucinated objects}\}\vert }{\vert\{\text{all objects mentioned}\}\vert }$$
$$\text{CHAIR}_{S} = \frac{ \vert\{ \text{sentences with hallucinated object}\}\vert }{\vert\{\text{all sentences}\}\vert }$$
$$\text{Recall} = \frac{ \vert\{ \text{correct mentioned objects}\}\vert }{\vert\{\text{ground truth objects}\}\vert }$$
}

To provide a more comprehensive assessment of hallucinations, we use the Sentence-level Hallucination Ratio (SHR)~\cite{zhao2023hallucinations}, a GPT-4-based evaluation metric. This metric includes hallucinations involving object existence, relationships, and attributes. We generate descriptions for 200 images from the VG dataset~\cite {krishna2016visualgenomeconnectinglanguage}, using the same prompts as in the CHAIR metric.
Specifically, SHR leverages GPT-4\footnote{We used GPT-4o (\texttt{gpt-4o-2024-08-06}) for hallucination judgement.} to compare the model's responses with the manually annotated descriptions from the VG dataset, evaluating each response on a sentence-by-sentence to identify potential hallucinations accurately.

\noindent\textbf{Baseline LVLMs.}
In LVLMs, two prominent methods for aligning text and vision modalities are the projection layer-based approach and the learnable query-based approach~\citep{li2023blip2bootstrappinglanguageimagepretraining,zhu2023minigpt4enhancingvisionlanguageunderstanding,chen2023shikraunleashingmultimodalllms,liu2023visualinstructiontuning}.
In our experiments, we utilize representative models for each aligning method: LLAVA-1.5 (7B/13B)~\cite{liu2024improvedbaselinesvisualinstruction} and InstructBLIP (7B/13B)~\cite{dai2023instructblipgeneralpurposevisionlanguagemodels}.

\noindent\textbf{Baseline Decoding Methods.}
We include various decoding methods as baseline approaches in our study, including greedy decoding, nucleus sampling, and beam search for traditional methods.
In addition, we incorporate various contrastive decoding techniques, including Visual Contrastive Decoding (VCD)~\cite{leng2023mitigatingobjecthallucinationslarge}, which contrasts the original image prompt with a distorted image prompt; Instruction Contrastive Decoding (ICD)~\cite{wang2024mitigatinghallucinationslargevisionlanguage}, which contrasts the original instruction prompt with a modified instruction prompt; and Multi-Modal Mutual Information Decoding (M3ID)~\cite{favero2024multimodalhallucinationcontrolvisual}, which contrasts the image prompt with a non-image prompt, with the contrast strength progressively increasing as the token length grows.
Lastly, we include OPERA~\cite{huang2024operaalleviatinghallucinationmultimodal}, a beam search-based method designed to counteract the model’s tendency to overemphasize specific anchor tokens.

\subsection{Main Results}
\label{main_result}
\noindent\textbf{Results on CHAIR evaluation.} 
As shown in Table~\ref{method_chair_table_2}, SumGD significantly outperforms the baseline methods in the CHAIR$_S$ and CHAIR$_I$ across different model sizes and architectures. 
Specifically, compared to Greedy decoding, SumGD-S achieves a 16.5\% improvement in CHAIR$_S$ and a 19\% improvement in CHAIR$_I$ on LLAVA 1.5 7B. On InstructBLIP 7B, the improvements are even more pronounced, with a 23.7\% improvement in CHAIR$_I$.

\begin{figure}[t]
  \centering
    \includegraphics[width=1\linewidth]{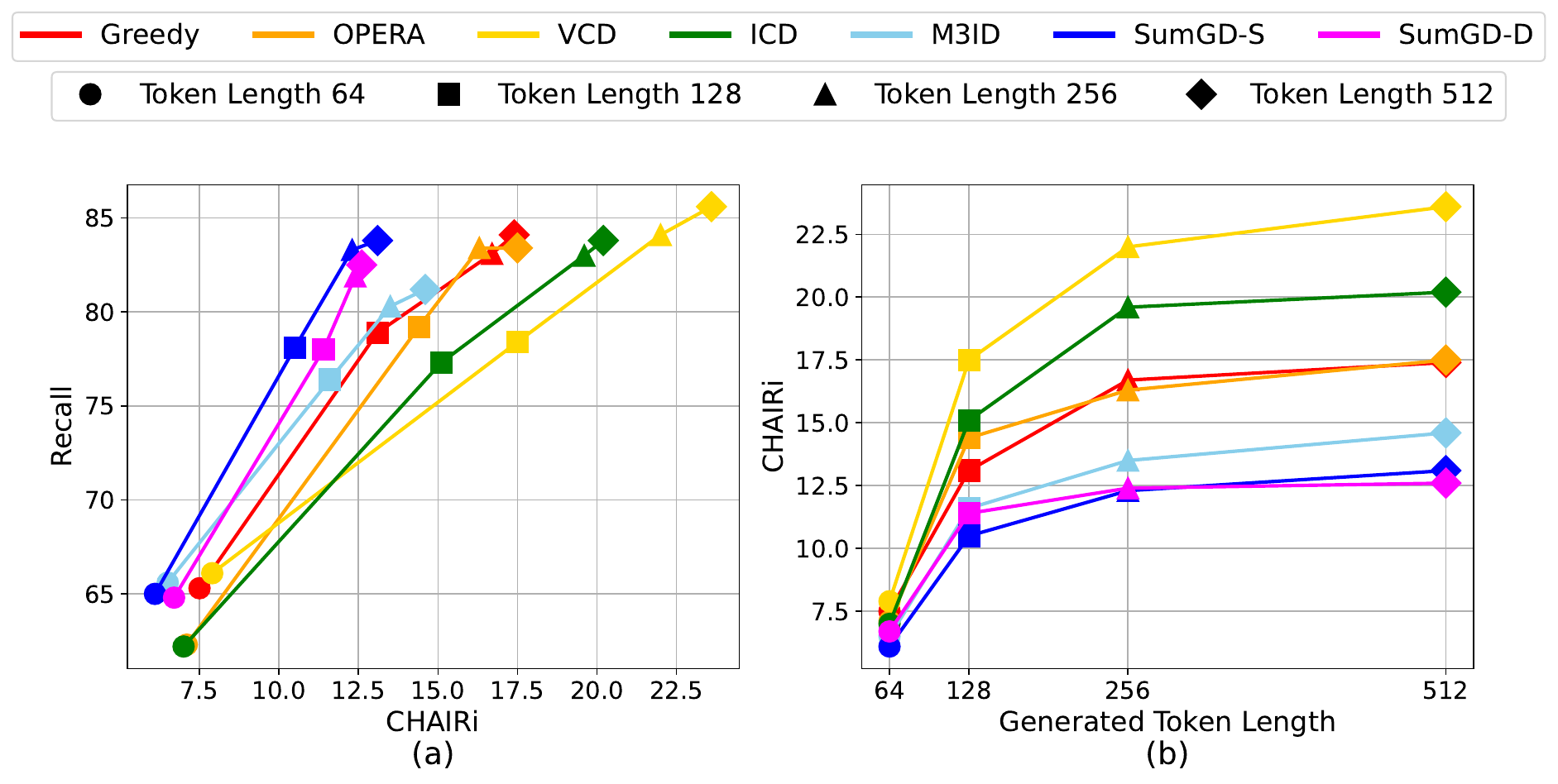}
    
\caption{
    \textbf{(Left)} A position closer to the top-left indicates an optimal balance between factuality and recall. \textbf{(Right)} Trade-off between generated token length and hallucination (lower is better).
    }
    \label{method_chair_figure}
\vspace{-4mm}
\end{figure}

We conduct the CHAIR evaluation by fixing the generated token lengths at 64, 128, 256, and 512, representing a range from short to long text generation to ensure a fair evaluation of object hallucination across different methods\footnote{CHAIR is a precision-based metric, which means it can be hacked by generating shorter captions or fewer objects.} (see Appendix~\ref{total_result_section} for full experimental results). 
As illustrated in Figure~\ref{method_chair_figure} (a)
our approach maintains a Pareto optimal position in the factuality-recall trade-off relative to all other methods. Notably, this robustness in managing the trade-off becomes more pronounced as the sequence length increases.
Furthermore, Figure~\ref{method_chair_figure} (b) shows that, even when considering object hallucination alone, our method exhibits the lowest degree of object hallucination across all variations of generated token lengths.
This result is significant, as it suggests that our method can capture both factual accuracy and detailed explanations across short and long generations. This demonstrates the broad applicability of our method.

\begin{table}[t]
\centering
\renewcommand{\arraystretch}{1}
\resizebox{0.7\linewidth}{!}{%
\begin{tabular}{lcccc}
\toprule
\multicolumn{1}{c}{\multirow{2}{*}{\raggedright Method}} & \multicolumn{2}{c}{LLAVA-1.5 7B} & \multicolumn{2}{c}{InstructBLIP 7B} \\
\cmidrule(r){2-3} \cmidrule(l){4-5}
 & SHR $\downarrow$ & SPI  & SHR $\downarrow$ & SPI  \\
\midrule
Greedy   & 43.3 & 5.00 & 47.4 & 5.14 \\ 
OPERA      & 42.0 & 4.74 & 46.4 & 4.76 \\ 
VCD      & 52.0 & 5.18 & 49.5 & 4.97 \\ 
ICD      & 50.2 & 4.93 & 57.8 & 5.93 \\ 
M3ID     & 46.4 & 5.02 & 59.9 & 5.51 \\ 
SumGD-D    & \underline{41.7} & 5.08 & \underline{46.1} & 5.26 \\ 
SumGD-S    & \textbf{40.8} & 5.03 & \textbf{45.7} & 5.30 \\ 
\bottomrule
\end{tabular}%
}
\caption{Results on Sentence-Hallucination Ratio (SHR) and Sentence Per Image (SPI) (\textit{max new tokens} is 512). The best performances within each setting are bolded, and the second-best are underlined.}
\label{tab:comparison}
\vspace{-5mm}
\end{table}

\noindent\textbf{Results on Sentence-level Hallucination Ratio.}
Table~\ref{tab:comparison} shows that SumGD-S achieves the lowest sentence-level hallucination rate on both the LLAVA 1.5 and InstructBLIP models. 
Additionally, SumGD-D ranks second on both models. Based on these results, our SumGD method demonstrates strong factual accuracy in holistic hallucination evaluations.
OPERA performs comparably to SumGD-D, but since it relies on beam search, it is less efficient than our method in terms of cost.
Moreover, an examination of the Sentences Per Image (SPI) reveals that our method does not achieve favorable results simply by generating fewer sentences.

\section{Analysis}
\subsection{Analysis of SumGD and Contrastive Decoding}
\label{analysis language prior text}

\begin{figure}[ht]
    \centering
    \includegraphics[width=0.8\linewidth]{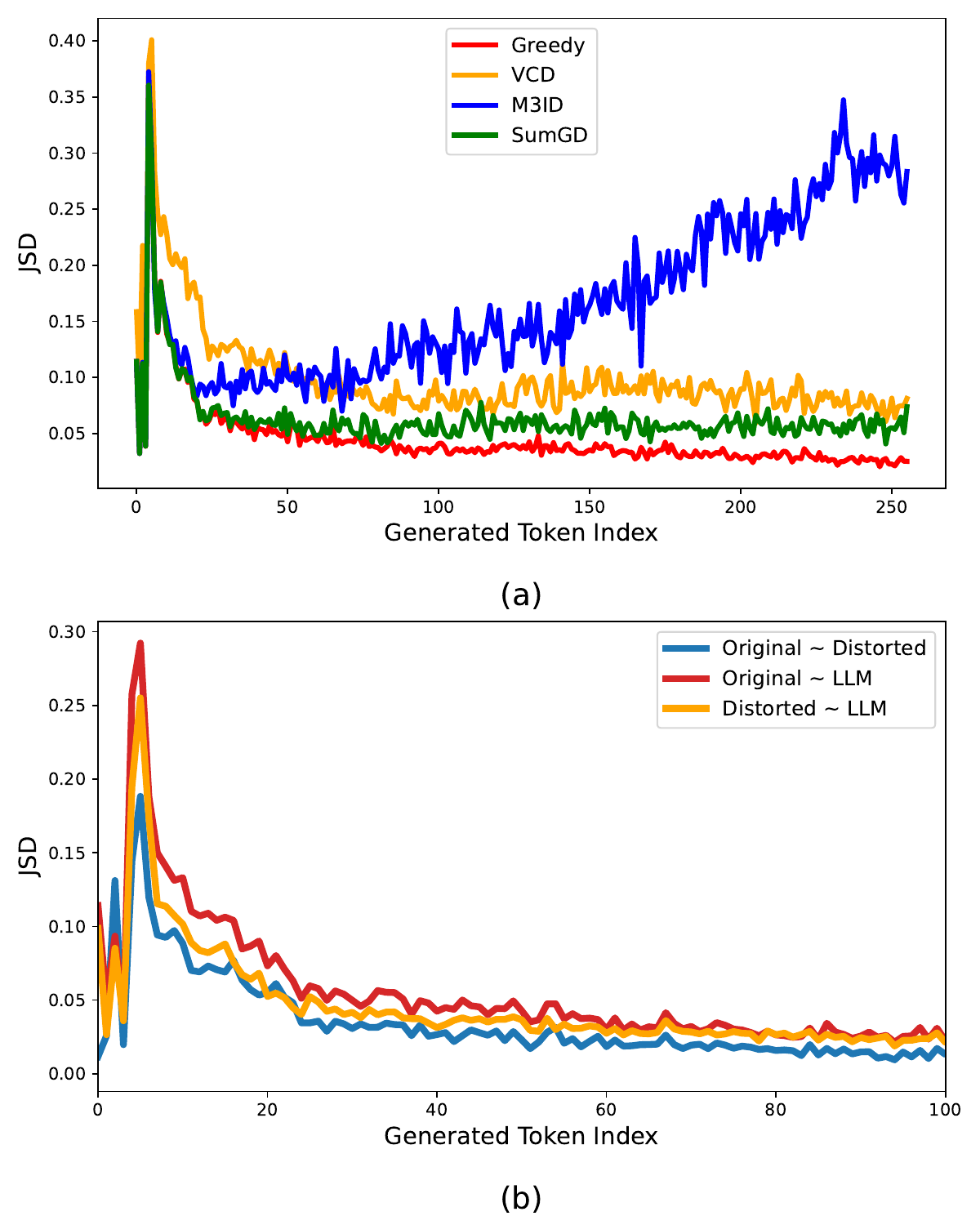}
    \caption{\textbf{(Top)} 
JSD between each method and LLM distributions at each decoding step. 
\textbf{(Bottom)} 
JSD between the Original Image and LLM, Distorted Image and LLM, Original Image and Distorted Image.}
    \label{section_5_first_figure}

\end{figure}

\begin{table}[ht]

    \resizebox{\linewidth}{!}{%
    \begin{tabular}{lcccccccc}
    \hline
    \multirow{2}{*}{Method} & \multicolumn{4}{c}{Token length 64} & \multicolumn{4}{c}{Token length 256} \\ \cmidrule(r){2-5} \cmidrule(l){6-9} 
                      & $C_s$ $\downarrow$     & $C_i$ $\downarrow$     & $R$ $\uparrow$ & $TQ$ $\uparrow$    & $C_s$ $\downarrow$    & $C_i$ $\downarrow$    & $R$ $\uparrow$  & $TQ$ $\uparrow$  \\ \hline
    Greedy            & 27     & 7.5   &65.3 & 4.97   & 67.5   & 16.7 &83.1  & 4.46   \\
    VCD               & 24.0   & 7.9  &66.1  & 4.93   & \textbf{82.5}     & \textbf{22.0}  &84.1 & 4.53   \\
    M3ID              & 20.5   & 6.5  &65.6  & \textbf{4.85}   & 62     & 13.5  &80.3 & \textbf{2.39}   \\
    SumGD               & 22.5   & 6.1  &65.0  & 4.93   & \textbf{54}     & \textbf{12.3}  &83.3 & \textbf{3.75}   \\ \hline
    \end{tabular}%
    }
    \caption{CHAIR metric and Text Quality in various generated token lengths. Denote CHAIR$_S$ as $C_S$, CHAIR$_I$ as $C_I$, Recall as $R$ and Text Quality as $TQ$.}
    \label{analysis prior table}\vspace{-4mm}

\end{table}
In this section, we analyze SumGD and contrastive decoding, focusing on their relationship with language priors. To explore this, we compute the JSD between each method's output and LLM distribution at each decoding step, followed by Section~\ref{how to measure JSD}. For the analysis, we generate descriptions for 200 images from the MSCOCO 2014 validation set using LLAVA 1.5 7B.
Factual accuracy is evaluated using the CHAIR metric, while text quality is assessed by GPT-4o~\cite{gpt4o} on a 1 to 5 scale (see details in Appendix~\ref{gpt-4o prompt}).

Two key questions guide the analysis. \textbf{Question1}: \textit{Is significantly deviating from language priors always beneficial?} \textbf{Question2}: \textit{Can contrastive decoding reduce hallucinations in LVLMs when language priors heavily influence the two output distributions being contrasted?}

To assess whether significantly deviating from language priors is always beneficial, we examine M3ID, a contrastive decoding method that progressively reduces language priors to focus more on visual information, as shown in Figure~\ref{section_5_first_figure} (a). However, as presented in Table~\ref{analysis prior table}, text quality drops considerably when generating up to 64 tokens compared to 256 tokens. Specifically, it declines from \textbf{4.85} to \textbf{2.39}, a reduction of about \textbf{50.7\%}.
This suggests that a significant deviation from the language prior disrupts the distribution of language-related tokens, leading to a degradation in text quality.

To investigate the effectiveness of contrastive decoding when language priors significantly influence the original distribution, we investigate VCD. In VCD, the output distribution of the original image prompt is contrasted with that of the distorted image prompt to produce outputs that more align with the original image.
A noteworthy observation is that both the output distributions of the original and distorted image prompts progressively converge towards the LLM distribution, as shown in Figure~\ref{section_5_first_figure} (b).
This finding indicates that language priors are influencing both the original output distribution and the output distribution that needs to be compared.
Consequently, the two distributions become increasingly similar, diminishing the effectiveness of contrastive decoding.
Table~\ref{analysis prior table} demonstrates the reduced effectiveness of contrastive decoding, as VCD results in more instances of object hallucination compared to greedy decoding.
Although current contrastive decoding methods focus on distorting the image to create meaningful differences from the original~\cite{leng2023mitigatingobjecthallucinationslarge,kim2024vacodevisualaugmentedcontrastive,wan2024contrastiveregionguidanceimproving}, the strong influence of language priors may obscure the intended effects of these distortions, undermining the effectiveness of contrastive decoding.
This finding is crucial to understanding the limitations of current contrastive decoding approaches.

\begin{figure}[t]
  \centering
    \includegraphics[trim=0 0 0 45, clip, width=0.9\linewidth]{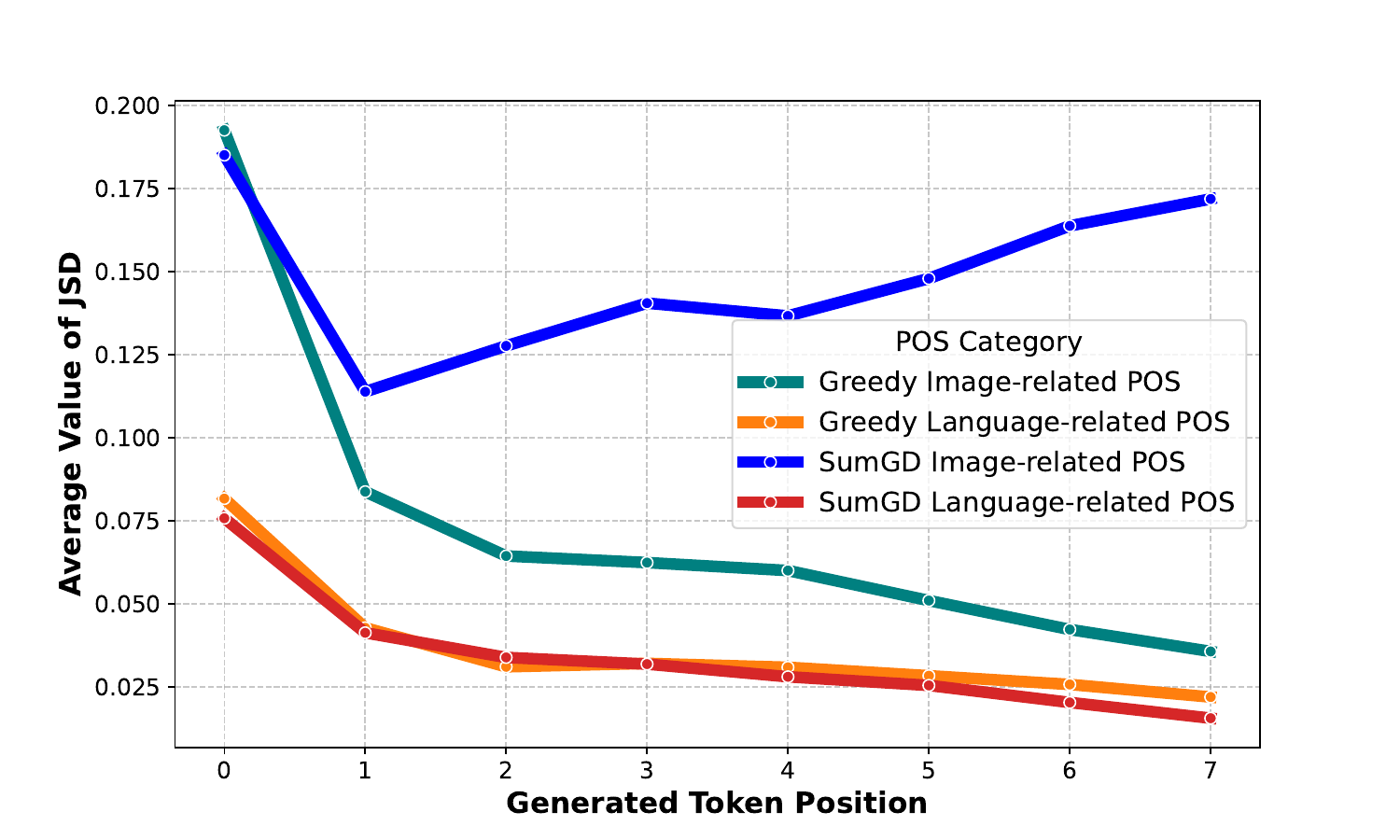}
\caption{
    Average JSD values for image-related POS and language-related POS across intervals of 32 tokens, measured in Greedy decoding and SumGD.
    }
    \label{effect_sum_fig}
\vspace{-4mm}
\end{figure}

Unlike contrastive decoding, SumGD excels at reducing object hallucinations while also maintaining a good balance in terms of text quality, as shown in Table~\ref{analysis prior table}. To further understand the effectiveness of SumGD, we measure how much SumGD and the Greedy method rely on language priors for each POS type. As seen in Figure~\ref{effect_sum_fig}, SumGD demonstrates a clear reduction in language priors when predicting image-related POS tokens, while preserving the original dependency on language-related POS tokens. These results indicate that our approach effectively mitigates language priors without compromising the core language modeling properties of the LVLM.

\begin{table*}[t]
\centering
\resizebox{0.8\linewidth}{!}{%
\begin{tabular}{lcccccc}
\hline
             & CHAIR$_S$ $\downarrow$ & CHAIR$_I$ $\downarrow$ & Recall $\uparrow$ & Text Quality $\uparrow$ & 1-gram $\uparrow$ & 2-gram $\uparrow$ \\ \hline
{\textit{\textbf{Baseline}}}       &    &    &    &           &   &   \\
Greedy Decoding       & 51.5   & 13.7   & 79.1   & 4.9          & 61.85  & 92.55  \\ \hline
{\textit{\textbf{Summary Models}}}       &    &    &    &           &   &   \\
Distilled-Flan-T5-base(248M)    & 42.5   & 11.8   & 77.8   & 4.8          & 59.87  & 90.57  \\
LLAVA 1.5(7B) & 43     & 11.1   & 79.1   & 4.84         & 60.63  & 91.41  \\
GPT-4o        & 43     & 10.3   & 78     & 4.77         & 59.36   & 89.77  \\ \hline
{\textit{\textbf{POS Control in SumGD}}}       &    &    &    &           &   &   \\
 ALL POS       & 39     & 10.1   & 75.8   & 4.06         & 52.67  & 80.66  \\
Image-related POS& 43     & 11.1   & \textbf{79.1}   & \textbf{4.84}         & \textbf{60.63}  & \textbf{91.41}  \\ \hline
\end{tabular}%
}
\caption{Ablation study in terms of Summary Quality and POS Control in SumGD (\textit{max new tokens} is 512).}
\label{tab:ablation}
\vspace{-3mm}
\end{table*}

\subsection{Ablation study of SumGD}
In this section, we conduct ablation studies to evaluate the quality of the summary and the effect of POS control in SumGD. For this, we use LLAVA 1.5 7B to generate descriptions for 200 images from the MSCOCO 2014 validation dataset.
We employ the CHAIR metric and the text quality metric as described in Section~\ref{analysis language prior text}. Additionally, we include the n-gram fluency indicator~\cite{zhao2024hallucinationsenhancinglvlmshallucinationaware}, represented by
$\frac{set(ngrams(s))}{len(ngram(s))}$, where $s$ denotes the description, to measure fluency.

\noindent\textbf{Summary Quality.}
We conduct an ablation experiment to evaluate the quality of summaries used in SumGD.
To achieve this, we employ three distinct summarization models—Distilled-Flan-T5-base, LLAVA 1.5 7B, and GPT-4o.
The results, as presented in Table~\ref{tab:ablation}, reveal that the effect of summarization quality is consistent across these models in terms of both CHAIR and text quality.
This suggests that both SumGD-D and SumGD-S achieve satisfactory levels of summarization quality.

\noindent\textbf{POS Control.}
We analyze the effect of applying image-related POS control in SumGD. As shown in Table~\ref{tab:ablation}, applying SumGD to all POS tokens, as well as selectively to image-related POS tokens, reduces object hallucination compared to the original decoding method.
However, when SumGD is applied to all POS tokens, text quality declines compared to the baseline, with the score dropping from \textbf{4.9} to \textbf{4.06}.
This decline is accompanied by a notable decrease in n-gram fluency and object recall, indicating more repetitive generation.
In contrast, when SumGD is applied only to image-related POS tokens, the resulting text quality remains almost unchanged, with the score only slightly decreasing from \textbf{4.9} to \textbf{4.84}.
These results demonstrate that applying SumGD selectively to image-related POS tokens effectively preserves the model's text quality.

\section{Related works}
\noindent\textbf{Mitigating Language Priors in LVLMs.} Large Vision-Language models (LVLMs) extend pre-trained Large Language Models (LLMs) by incorporating visual tokens, enabling them to process visual content~\cite{liu2023visualinstructiontuning,dai2023instructblipgeneralpurposevisionlanguagemodels,zhu2023minigpt4enhancingvisionlanguageunderstanding}. In LVLM architectures, the language model is significantly larger than the vision model, creating an imbalanced structure where the language model exerts more significant influence. As a result of this imbalance, the model tends to rely on linguistic patterns rather than adequately considering the visual information provided, a phenomenon known as the language prior problem~\cite{guan2024hallusionbenchadvanceddiagnosticsuite,lee2024vlindbenchmeasuringlanguagepriors,lee-etal-2024-volcano}.
Several studies have explored ways to control LLM outputs to better align with desired objectives~\cite{li-etal-2023-contrastive,hallinan-etal-2023-detoxifying,kim-etal-2023-critic,kim2024guaranteedgenerationlargelanguage}. Similarly, research on LVLMs has focused on contrastive decoding techniques to reduce the model's over-reliance on language priors~\cite{manevich-tsarfaty-2024-mitigating}.
Visual Contrastive Decoding (VCD)~\cite{leng2023mitigatingobjecthallucinationslarge} works by utilizing distorted images, which amplify the language prior, and Instruction Contrastive Decoding (ICD)~\cite{wang2024mitigatinghallucinationslargevisionlanguage} introduces misleading instructions to achieve a similar effect. Both methods aim to reduce the language prior’s dominance by leveraging these amplified conditions to adjust the model's behavior. Additionally, Multi-Modal Mutual Information Decoding (M3ID)~\cite{favero2024multimodalhallucinationcontrolvisual} identified that as the token length increases, the model dilutes visual information, leading to a more substantial reliance on language priors. To counter this, M3ID applies more assertive contrastive decoding techniques as the token length grows to calibrate the model's over-reliance on language priors.
However, contrastive decoding can disrupt the distribution of tokens essential for language modeling, leading to a decline in text quality.
Additionally, due to the language prior, the two output distributions being contrasted may become more similar, making it less effective in reducing hallucinations.
Our method, Summary-Guided Decoding (SumGD), addresses these issues by using summarization techniques to naturally reduces the influence of language priors, allowing the model to focus more on the image. Furthermore, SumGD preserves text quality by controlling only the POS tokens relevant to the image.


\section{Conclusion}

In this paper, we introduce Summary-Guided Decoding (SumGD) as a novel method to mitigate object hallucinations in LVLMs.
Our analysis reveals that as token sequences grow, LVLMs tend to increasingly rely on language priors, reducing the influence of visual information during the decoding process. To address this, SumGD employs summarization techniques to shorten token length, encouraging the model to incorporate more visual details while controlling only the image-related POS tokens to maintain text quality.
Our experimental results demonstrate that SumGD significantly reduces object hallucination and achieves an optimal balance between factual accuracy and recall in both short and long description tasks.

\section*{Limitations}
In this paper, we propose a Summary-Guided Decoding (SumGD) to mitigate object hallucinations in Large Vision-Language Models (LVLMs).
However, this approach comes with some limitations.

First, the generation of summaries during the decoding process incurs additional computational cost, resulting in increased inference time.

Second, while summarization effectively reduces input length and helps mitigate hallucinations, it may also result in the loss of critical contextual information. Additionally, excessively long summaries can increase LVLMs' dependence on language priors, which may degrade the performance of SumGD. Therefore, it is crucial for future work to train LVLMs in a way that inherently avoids over-reliance on language priors, even when token lengths are extended.

Lastly, we employ part-of-speech (POS) tagging to distinguish between image-related and language-related tokens. However, relying solely on POS tagging for this differentiation can be problematic. The development of more advanced methods for token distinction could enhance the effectiveness of SumGD and create further synergies with this approach.

\section*{Ethics Statement}
In this paper, we contribute to the future development of a safe and reliable AI community by conducting research focused on reducing hallucinations in Large Vision-Language Models~\cite{xie-etal-2024-v,manevich-tsarfaty-2024-mitigating}.

Our experiments were conducted by using publicly available datasets, ensuring that no private or sensitive personal data was involved.
Furthermore, we utilized publicly accessible models for our experiments, reinforcing the transparency and reproducibility of our approach.

However, the models we used may still exhibit biases inherent in the underlying datasets and training processes~\citep{howard2024uncoveringbiaslargevisionlanguage,kim-etal-2024-lifetox,fraser-kiritchenko-2024-examining,lee2024llmjudgesrobustexpressionsuncertainty,koh-etal-2024-llms}. While our focus was on biases related to language priors, we acknowledge the need to address other potential biases as well.

\section*{Acknowledgments}

This work was partly supported by the Institute of Information \& Communications Technology Planning \& Evaluation(IITP)-ITRC(Information Technology Research Center) grant funded by the Korea government(MSIT)(IITP-2025-RS-2024-00437633, 60\%), Institute of Information \& communications Technology Planning \& Evaluation (IITP) grant funded by the Korea government(MSIT) [NO.RS-2021-II211343, Artificial Intelligence Graduate School Program (Seoul National University)] \& RS-2021-II212068, Artificial Intelligence Innovation Hub (Artificial Intelligence Institute, Seoul National University)], and the BK21 FOUR program of the Education and Research Program for Future ICT Pioneers, Seoul National University in 2024.
K. Jung is with ASRI, Seoul National University, Korea. The Institute of Engineering Research at Seoul National University provided research facilities for this work.


\bibliography{custom}

\clearpage
\appendix


\section{Details of Code, Hyperparameters, and GPU Cost}
\label{reproduction}
We conduct our experiments based on the OPERA~\cite{huang2024operaalleviatinghallucinationmultimodal} code base which is publicly available.
We use the publicly available code provided by the authors for the VCD and OPERA methods.
While M3ID and ICD are implemented from scratch due to the lack of public code. For VCD, OPERA, and ICD, we use the hyperparameters as specified in their respective papers. Since only LLAVA 1.5's hyperparameters were reported in M3ID, we apply these hyperparameters to both LLAVA 1.5 and InstructBLIP for our experiments. Also, we set repetition penalty as 1.
All the experiments are conducted using 1 NVIDIA RTX A6000 GPU.

\section{Experimental Settings for Analyzing Language Priors}
\label{motivation_experiemtn_settings}
We generate descriptions using LLAVA 1.5 7B for 5000 images from the MSCOCO 2014 validation dataset~\cite{lin2015microsoftcococommonobjects} and annotate each token to determine whether it represents an object hallucination. An object hallucination is defined as an object not present in the image. We employ the CHAIR metric pipeline~\cite{rohrbach2019objecthallucinationimagecaptioning} for evaluation.

\section{Ablation Study of Inference Time Cost}
\label{time cost}

\begin{table}[ht]
\centering
\resizebox{\linewidth}{!}{%
\begin{tabular}{lcccc}
\hline
\textbf{Method} & \textbf{$RIC$} & \textbf{$C_s$} & \textbf{$C_i$} & \textbf{$R$} \\
\hline
Greedy & 1 & 51.5 & 13.7 & 79.1 \\
VCD & 2 & 58.0 & 16.4 & 77.8 \\
Beam Search & 5 & 47.5 & 12.5 & 79.2 \\
OPERA & 5$\uparrow$ & 46.0 & 13.4 & 78.3 \\
SumGD-S &  &  &  &  \\
\quad +Summarization & 2.54 & - & - & - \\
\quad +Summarization + POS Tagging & 2.98 & 43.5 & 11.0 & 79.2 \\
SumGD-D &  &  &  &  \\
\quad +Summarization & 1.87 & - & - & - \\
\quad +Summarization + POS Tagging & 2.3 & 42.5 & 11.6 & 77.7 \\
\hline
\end{tabular}%
}
\caption{Comparison of Methods with Relative Inference Costs and CHAIR Metrics. Denote CHAIR$_S$ as $C_S$, CHAIR$_I$ as $C_I$, Recall as $R$ and Relative Inference Costs as $RIC$.}
\label{table:inference_cost_comparison}
\end{table}
We conduct an ablation study to evaluate the impact of the summarization process on inference time. Since the summarization process inherently requires additional token generation, which affects efficiency, we measure inference costs (normalized by relative token generation costs) across baseline methods using the LLAVA 1.5 7B model. Our results show that the distilled summary model (Flan-T5-base) in SumGD-D requires only half the time per token compared to the LLAVA 1.5 7B model. This finding is incorporated into our inference cost calculations. Additionally, to ensure accurate part-of-speech (POS) tagging, we generate one extra word after each current token and then measure the current token’s POS tag. We perform this ablation study on 200 images. The comparison of inference time costs is presented in Table~\ref{table:inference_cost_comparison}.

\section{Experiments on LLaVA 1.6}
\label{llava-next}
\begin{table}[ht]
\centering
\resizebox{\linewidth}{!}{
\begin{tabular}{lcc}
\hline
\textbf{Method} & \textbf{CHAIR$_S$} & \textbf{CHAIR$_I$} \\
\hline
Greedy & 36.5 & 10.2 \\
Nucleus & 37.0 & 9.7 \\
Beam-Search (beams=5) & 34.5 & 10.5 \\
SumGD-S & 30.5 & 6.4 \\
\hline
\end{tabular}
}
\caption{CHAIR results on LLAVA 1.6 7B (\textit{max new tokens} is 512).}
\label{table:llava-1.6-chair}
\end{table}
We conduct an evaluation of our SumGD method on the latest model, LLaVA 1.6 7B model~\cite{liu2024llavanext}, to assess its effectiveness. Specifically, we perform the CHAIR evaluation (lower is better) on 200 images. Table~\ref{table:llava-1.6-chair} shows that SumGD-S effectively reduces object hallucination, demonstrating the applicability of our methodology even with the latest models.

\section{Distilled Flan-T5-base model}
\label{distill_training}
We employ LLAVA 1.5 7B to perform Summary-Guided Decoding with Self-Summarization while generating descriptions for 5,000 images from the MSCOCO dataset. During this process, LLAVA 1.5 iteratively summarizes each previous sentence, and we store each previous sentence along with its corresponding summarized sentence as a pair. This paired dataset is subsequently used to fine-tune the Flan-T5-base model with the prompt ``What is a summary of this text?'' for training purposes.

\section{Summarize Prompt for Summary-Guided Decoding}
In SumGD-S, we use summary prompt as:
\begin{verbatim}
USER: Summarize the following 
caption in briefly.
\nCaption: <<caption>> ASSISTANT:
\end{verbatim}
In SumGD-D, we use summary prompt as:
\begin{verbatim}
<<Caption>> \nWhat is a summary of 
this text?
\end{verbatim}

\section{GPT-4o Prompt for text quality evaluation}

\label{gpt-4o prompt}

\begin{figure}[H]
    \centering
    \small
    \begin{tcolorbox}
    [width=\linewidth, sharp corners=all, colback=gray!10, boxrule=0.3mm]
    \#\#\#Task Description: \\
You will be given one caption written for a given image.
Your task is to rate the caption on one metric.
Please make sure you read and understand these instructions carefully. Please keep this document open while reviewing, and refer to it as needed.
The output format should look as follows: Score: [RESULT] (an integer number between 1 and 5).
Please do not generate any other opening, closing, and explanations.
\\ \\
\#\#\#Evaluation Criteria:\\
Text Quality (1-5) - Evaluate how well-written the caption is. A high-quality caption is clear, concise, grammatically correct, and well-structured.
\\ \\
\#\#\#Evaluation Steps:\\
1. Read the caption carefully and evaluate its clarity, grammar, and overall readability.\\
2. Check for any awkward phrasing, grammatical errors, or unnecessary complexity.\\
3. Assign a score for text quality on a scale of 1 to 5, where 1 is the lowest and 5 is the highest based on the Evaluation Criteria.
\\ \\
\#\#\#Given Caption:\\
\{\{Caption\}\}
\\ \\
\#\#\#Score:
    \end{tcolorbox}
    \vspace{-7pt}
    \caption{GPT-4o prompt for text quality evaluation}
    \label{fig:sample_prompt}
    \vspace{-10pt}
\end{figure}






\section{CHAIR metric on various token length}
\label{total_result_section}
In this section, we report CHAIR metric based on various generated token length.

\begin{table}[H]
\centering
\resizebox{0.9\linewidth}{!}{
\begin{tabular}{ccccc}
\hline
\textbf{Token Length} & \textbf{Method} & \textbf{CHAIRs} & \textbf{CHAIRi} & \textbf{Recall} \\
\hline
64  & Greedy             & 27   & 7.5  & 65.3 \\
64  & Nucleus            & 31.5 & 9.8  & 58.9 \\
64  & Beam               & 20   & 5.9  & 62.5 \\
64  & VCD                & 24.0 & 7.9  & 66.1 \\
64  & ICD                & 21.5   & 7.0  & 62.2 \\
64  & M3ID               & 20.5 & 6.5  & 65.6 \\
64  & OPERA              & 22.5   & 7.1  & 62.3 \\
64  & SumGD-S       & 22.5 & 6.1  & 65.0 \\
64  & SumGD-D  & 24   & 6.7  & 64.8 \\
\hline
128 & Greedy             & 53   & 13.1 & 78.9 \\
128 & Nucleus            & 56.5 & 16.5 & 74.2 \\
128 & Beam               & 50.5 & 13.3 & 78.3 \\
128 & VCD                & 63.0   & 17.5 & 78.4 \\
128 & ICD                & 56.0   & 15.1 & 77.3 \\
128 & M3ID               & 46.5 & 11.6 & 76.4 \\
128 & OPERA              & 49.5 & 14.4 & 79.2 \\
128 & SumGD-S       & 43.5 & 10.5 & 78.1 \\
128 & SumGD-D  & 43.5 & 11.4 & 78.0 \\
\hline
256 & Greedy             & 67.5 & 16.7 & 83.1 \\
256 & Nucleus            & 78 & 20.9 & 82.8 \\
256 & Beam               & 70   & 16.2 & 81.6 \\
256 & VCD                & 82.5   & 22.0 & 84.1 \\
256 & ICD                & 71   & 19.6 & 83.0 \\
256 & M3ID               & 62   & 13.5 & 80.3 \\
256 & OPERA              & 64.5   & 16.3 & 83.4 \\
256 & SumGD-S       & 54   & 12.3 & 83.3 \\
256 & SumGD-D  & 56.5 & 12.4 & 81.9 \\
\hline
512 & Greedy             & 69.5 & 17.4 & 84.1 \\
512 & Nucleus            & 80   & 22.0 & 83.8 \\
512 & Beam               & 71.5 & 17.4 & 82.3 \\
512 & VCD                & 83.0 & 23.6 & 85.6 \\
512 & ICD                & 73.0 & 20.2 & 83.8 \\
512 & M3ID               & 65.5 & 14.6 & 81.2 \\
512 & OPERA              & 66.5 & 17.5 & 83.4 \\
512 & SumGD-S       & 59   & 13.1 & 83.8 \\
512 & SumGD-D  & 61.5 & 12.6 & 82.5 \\
\hline
\end{tabular}
}
\caption{Performance comparison for CHAIRs, CHAIRi, and Recall}
\label{appendix_variation_results}
\end{table}

 

\section{Case Study}
 
This is the case study of Summary-Guided Decoding (SumGD), Visual Contrastive Decoding (VCD), and Multi-Modal Mutual Information Decoding (M3ID) in generating up to 256 tokens in detailed captioning task.
Case study shows that SumGD generated image-related words, while VCD showed a tendency to hallucinate by relying on the word `tie' during decoding. Additionally, M3ID exhibited issues in language modeling.

\begin{figure*}[t]
\centering
\includegraphics[width=1.0\textwidth]{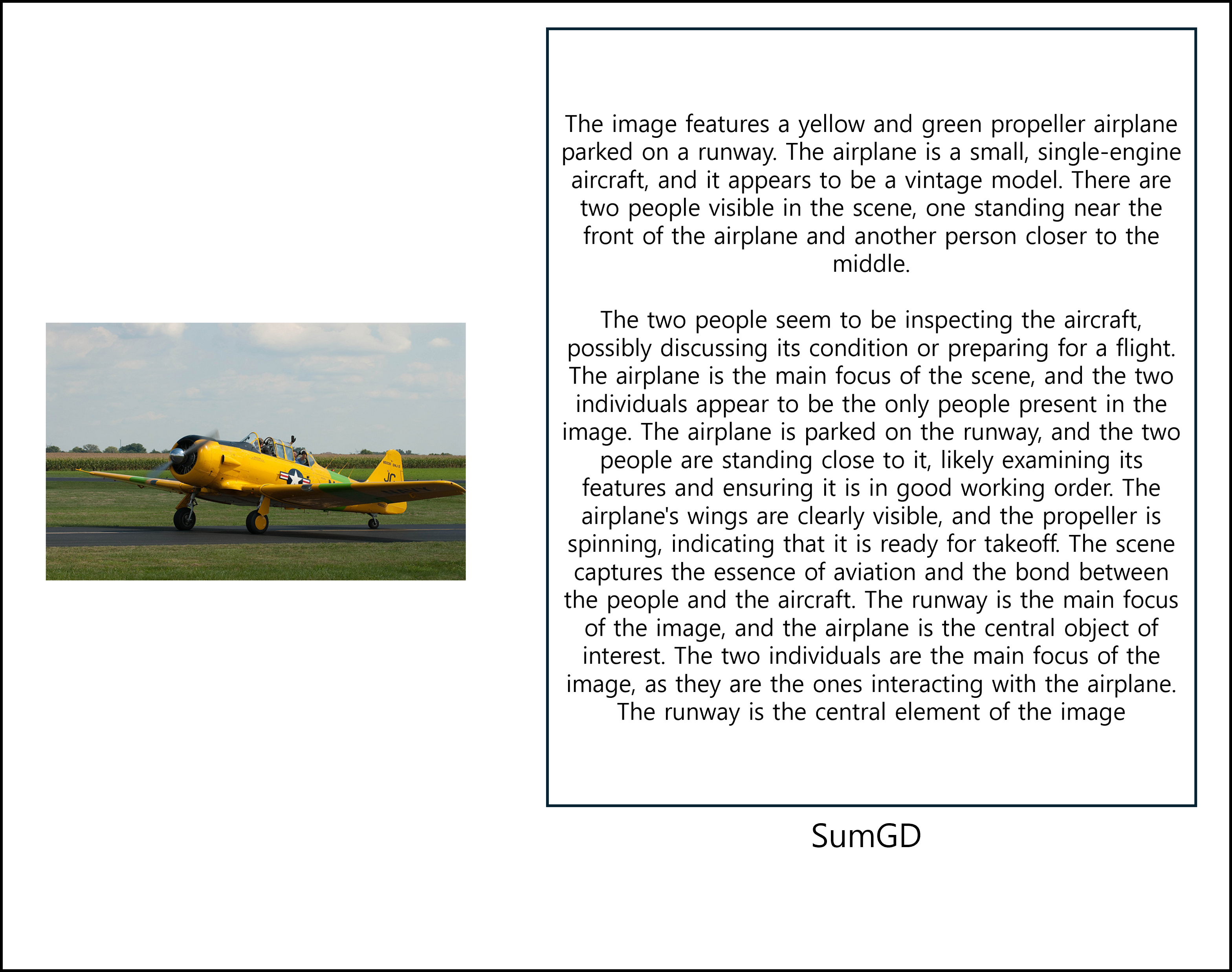} 
\caption{SumGD case study.}
\end{figure*}

\begin{figure*}[t]
\centering
\includegraphics[width=1.0\textwidth]{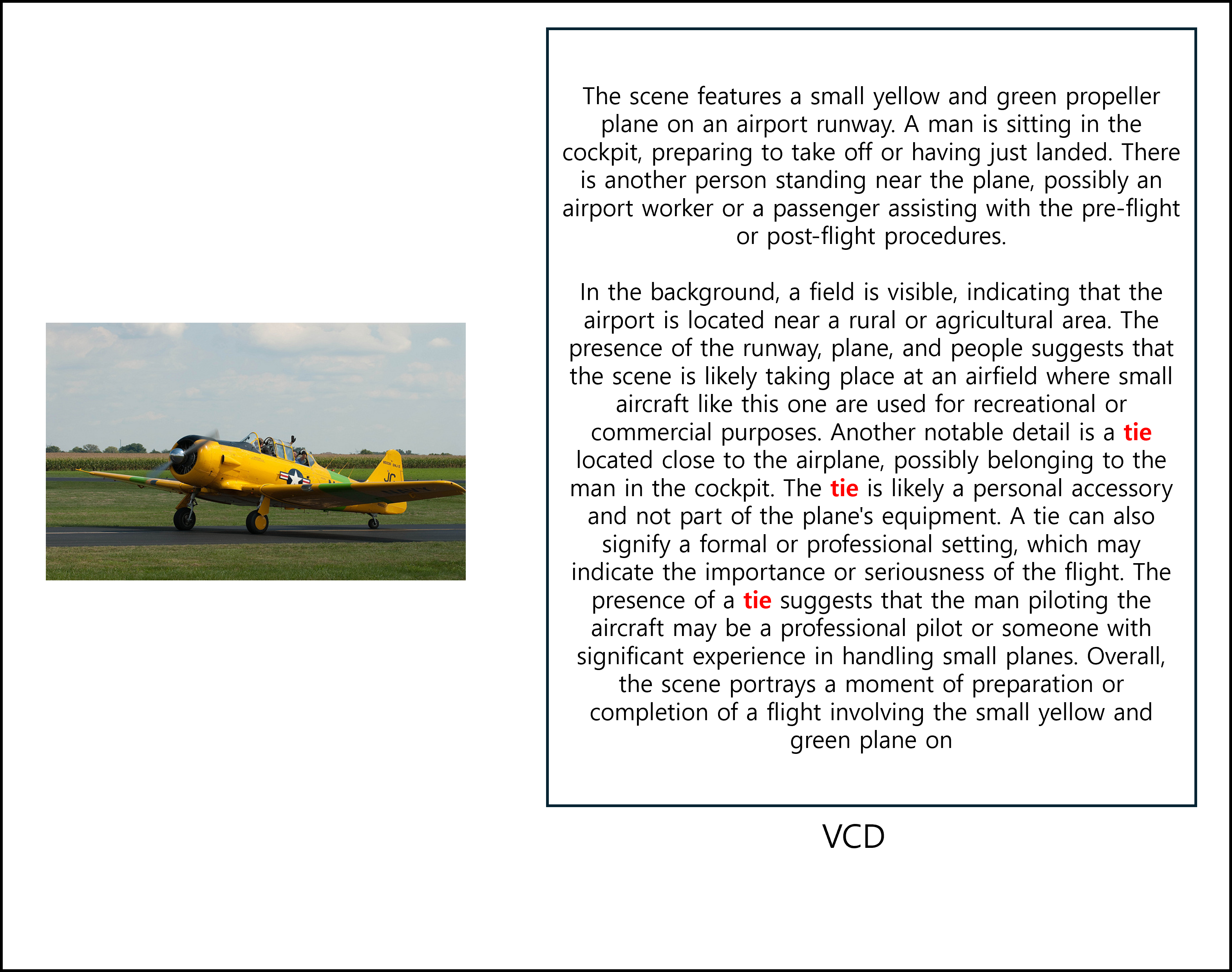} 
\caption{VCD case study. VCD heavily relies on word "tie" to generate descriptions which is not in the provided image.}
\end{figure*}

\begin{figure*}[t]
\centering
\includegraphics[width=1.0\textwidth]{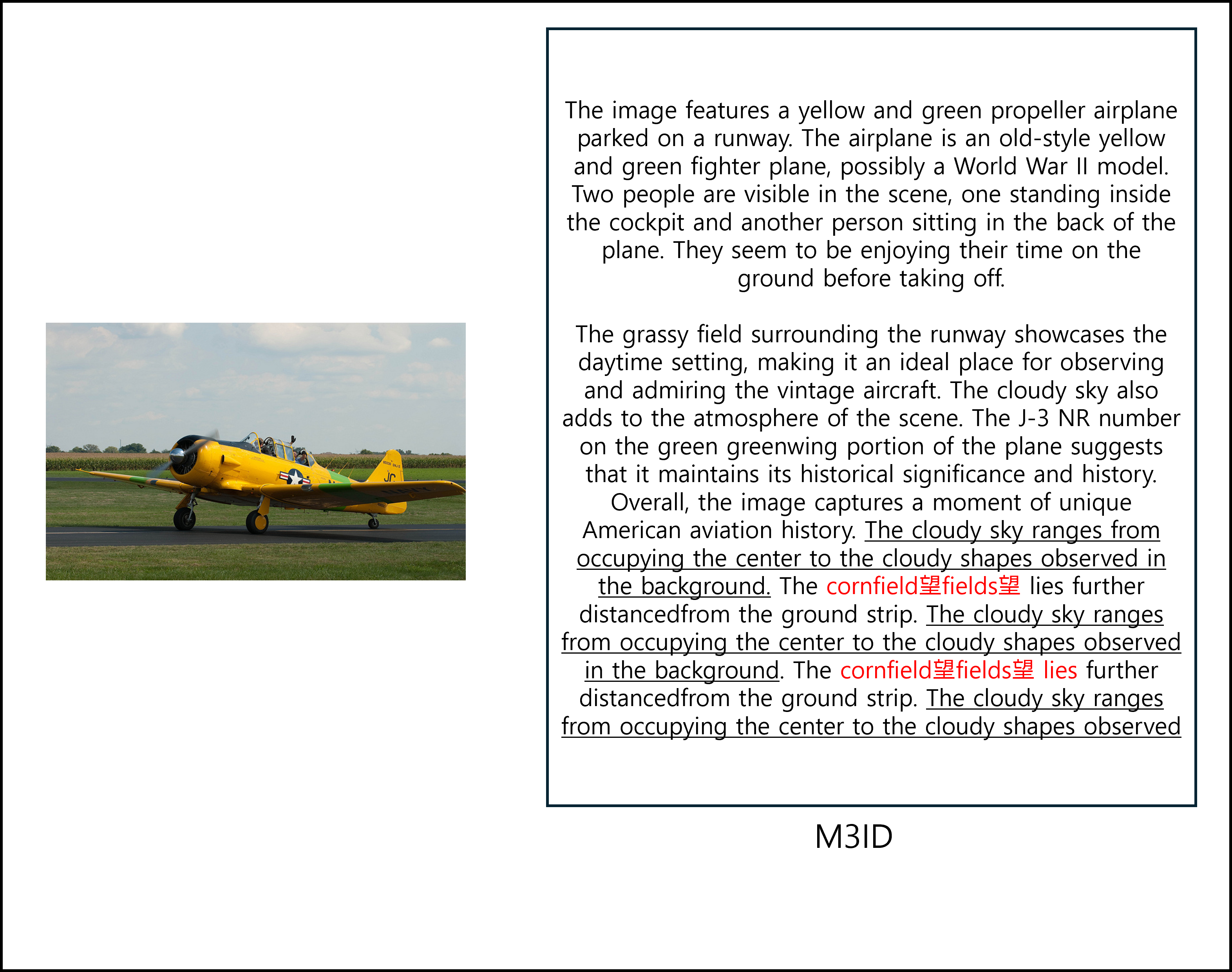} 
\caption{M3ID case study. Underline is for repetitive sentences. Red font denotes a degradation of language modeling.}
\end{figure*}

\end{document}